\documentclass[11pt,a4paper]{article}
\usepackage{paclic32}
\usepackage{times}
\usepackage{latexsym}
\usepackage{mycommand}
\usepackage[pdftex]{graphicx}
\usepackage{subfig}
\usepackage[hyphens]{url}

\usepackage[whole]{bxcjkjatype}
\usepackage{booktabs}
\usepackage{multirow}

\title{Japanese Sentiment Classification \\ using a Tree-Structured Long Short-Term Memory with Attention}

\author{
  Ryosuke Miyazaki\thanks{~~Now at Yahoo Japan Corporation.} \\
  Graduate School of System Design \\
  Tokyo Metropolitan University \\
  {\tt tmcit.miyazaki@gmail.com} \\\And
  Mamoru Komachi \\
  Graduate School of System Design \\
  Tokyo Metropolitan University \\
  {\tt komachi@tmu.ac.jp} \\
}

\date{}

\begin{document}
\maketitle

\begin{abstract}
Previous approaches to training syntax-based sentiment classification models required phrase-level annotated corpora, which are not readily available in many languages other than English.
Thus, we propose the use of tree-structured Long Short-Term Memory with an attention mechanism that pays attention to each subtree of a parse tree.
Experimental results indicate that our model achieves state-of-the-art performance for a Japanese sentiment classification task.
\end{abstract}

\section{Introduction}
Traditional approaches for sentiment classification rely on simple lexical
features, such as a bag-of-words, that are ineffective for many sentiment classification tasks \cite{Pang:2002}.
For example, the sentence ``Insecticides kill pests.'' contains both {\it
kill} and {\it pests}, indicating negative polarity.
But, the overall expression is still deemed positive.

To address this problem of polarity shift, \newcite{Nakagawa:2010} presented a
dependency-tree-based approach for the sentiment classification of a
sentence.
Their method assigns sentiment polarity to each subtree as a hidden variable that is not observable in the training data.
The polarity of the overall sentence is then classified by a tree-conditional
random field (Tree-CRF), marginalizing over the hidden variables representing
the polarities of the respective subtrees.
In this manner, the model can handle polarity-shifting operations such as negation.
However, this method suffers from feature sparseness because almost all features are combination features.

To overcome the data sparseness problem, deep-neural-network-based methods have attracted much attention because of their ability to use dense feature representations \cite{Socher:2011,Socher:2013,Kim:2014,Kalchbrenner:2014,Tai:2015,Zhang:2015}.
In particular, tree-structured approaches called recursive neural networks (RvNNs) have been shown to perform well in sentiment classification tasks \cite{Socher:2011,Socher:2013,Kim:2014,Tai:2015}.
Whereas Tree-CRF employs sparse and binary feature representations, RvNNs avoid feature sparseness by learning dense and continuous feature representations.
However, annotation for each phrase is crucial for learning RvNN models,
but there is no phrase-level annotated corpus in any language other than
English.

We therefore propose an RvNN model with an attention mechanism 
and augment the training example with polar dictionaries 
to compensate for the lack of phrase-level annotation.
Although \newcite{Kokkinos:2017} also provided an attention mechanism for
phrase-level annotated corpus, our model performs well on a sentence-level
annotated corpus through the introduction of polar dictionaries.

The main contributions of this work are as follows.
\begin{itemize}
\item We show that RvNN models can be learned from a sentence-level
    polarity-tagged Japanese corpus using an attention mechanism
    and polar dictionaries. 
\item We achieve the state-of-the-art performance in a Japanese sentiment classification task.
\item We have released our code on
GitHub.\footnote{See \url{https://github.com/tmu-nlp/AttnTreeLSTM4SentimentClassification}}
\end{itemize}

The rest of this paper is organized as follows.
Section \ref{sect:relatedwork} introduces related work.
Section \ref{sect:method}
describes our proposed method using Tree-LSTM with an attention mechanism and polar
dictionaries. Section \ref{sect:experiments} presents the experimental results
from Japanese and English sentiment datasets, and Section \ref{sect:discussion}
discusses the advantages and disadvantages of the proposed method.
Finally, Section \ref{sect:conclusion} concludes our work.

\section{Related Work}
\label{sect:relatedwork}

This section describes related work on Japanese sentiment classification, RvNNs, and attentional models.

\subsection{Japanese Sentiment Analysis}

\newcite{Nakagawa:2010} proposed a dependency-based polarity classifier.
Their model infers polarity from the composed nodes of a dependency tree using Tree-CRF.
Each subtree is represented as a hidden variable in consideration of interactions between the hidden variables.
A polar dictionary is used as the initial variable, and a polarity-reversing
word dictionary is used to capture whether the constructed phrase polarity is
reversed or not.
Our model uses only a polar dictionary and attempts to learn polarity shifting via RvNNs.

\newcite{Zhang:2015} adopted a stacked denoising auto-encoder.
Their model treats an input sentence as an average vector of their word
vectors, which is then fed into a stacked denoising auto-encoder.
Although this model omits syntactic information, it achieves high performance through its generalization ability.
However, it is not straightforward to employ polar dictionaries in their model.

\subsection{Recursive Neural Networks}
There are various RvNN models for sentiment classification \cite{Socher:2011,Socher:2012,Socher:2013,Qian:2015,Tai:2015,Zhu:2015}.
All of these models attempt to capture sentence representation in a bottom-up fashion in accordance with a parse tree.
In this way, sentence representation can be calculated by learning compositional functions for each phrase.

Several studies have focused on using a compositional function to improve compositionality \cite{Socher:2012,Socher:2013,Qian:2015,Tai:2015}.
\newcite{Socher:2012} parameterized each word as a matrix--vector combination to denote modification and representation.
\newcite{Socher:2013} used a bilinear function enacted by tensor slicing for composition in place of a large matrix--vector parameterization.
\newcite{Qian:2015} incorporated a constituent label of each phrase as feature embedding to take account of the different compositionality based on parent or children label combinations.
\newcite{Tai:2015} proposed a more robust model than those discussed above in which long short-term memory (LSTM) units were applied to a RvNN, as mentioned in Section \ref{sect_treelstm}.
Thanks to the application of LSTM, this model can learn increased numbers of
parameters appropriately, unlike the other models.

\subsection{Attentional Models}
Based on psychological studies, the human ability of intuition of attention \cite{Rensink:2000} has been introduced into many computer science fields.
The main function of this ability is deciding which part of an input needs to be focused on.

In natural language processing (NLP), the attention mechanism is utilized for many tasks, including neural machine translation \cite{Bahdanau:2015,Luong:2015,Eriguchi:2016}, neural summarization \cite{Hermann:2015,Rush:2015,Vinyals:2015}, representation learning \cite{Ling:2015}, and image captioning \cite{Xu:2015}.

These approaches incorporate consideration as to how each source word or region contributes to the generation of a target word.
Unlike most of the above-mentioned NLP tasks for generating word sequences, our model derives attention information for sentiment classification.
\newcite{Eriguchi:2016} proposed an attentional model focusing on phrase structure; our model omits the word-level recurrent LSTM layer from their model.
\newcite{Yang:2016} presented a hierarchical attention network for document
classification, incorporating sentence- and word-level attention mechanisms.
Our task addresses sentence-level sentiment classification, so that our model
employs a phrase-level (constituent) attention network rather than
sentence-level information in a document.

Probably the most related work to our study is \cite{Kokkinos:2017} 
and \cite{Zou:2018}. \newcite{Kokkinos:2017} built a similar model to ours,
but they did not use the hidden state of a RvNN in their final softmax, and
they also did not emphasize the use of polar dictionaries.
\newcite{Zou:2018} proposed a lexicon-based supervised attention model to
take advantage of a sentiment lexicon. They injected type-level lexical
information into an additional attention network, whereas we injected
token-level lexical information into a single RvNN model as a phrase-level
annotation.

\section{Attentional Tree-LSTM}
\label{sect:method}

\subsection{Tree-Structured LSTM}
\label{sect_treelstm}
Various RvNN models for handling sentence representation considering syntactic structure have been studied \cite{Socher:2011,Socher:2012,Socher:2013,Qian:2015,Tai:2015,Zhu:2015}.
RvNNs construct a sentence representation from their phrase representations by applying a composition function.
Phrase representations can be calculated by recursively adopting composition functions.
Binarizations of parse trees are often used to simplify the composition function.
In a parse tree, the root node, non-terminal node, and terminal node represent sentence, phrase, and word representations, respectively.

The $i$th non-terminal node representation $h_i$ is calculated by using the composition function $g$ as
\begin{eqnarray}
    h_i & = & f(g(h_i^l, h_i^r)), \\
    g(h_i^l, h_i^r) & = & W \coltwovec{h_i^l}{h_i^r} + b,
\end{eqnarray}
where the matrix $W \in R^{d \times 2d}$ and the bias $b \in R^d$ are the parameters to be learned, $h_i^l, h_i^r \in R^d$ are $d$-dimensional children vectors of node $h_i$, and the resulting vector $h_i$ is another $d$-dimensional vector.
The hyperbolic tangent is usually employed as the activation function $f$.
These RvNN models are essentially identical to recurrent neural models in that they are not able to retain a long history.

\newcite{Tai:2015} addressed this problem by introducing LSTM
\cite{Hochreiter:1997} to make RvNN less prone to the exploding/vanishing gradient problem.
In this paper, we use the Binary Tree-LSTM proposed by \newcite{Tai:2015} as an example of a tree-structured LSTM.
The Binary Tree-LSTM composes children vectors using the following equations:
\begin{eqnarray}
    i_j & = & \sigma\left(U^{(i)} \coltwovec{h_j^l}{h_j^r} + b^{(i)}\right), \\
    f_{jl} & = & \sigma\left(U^{(fl)} h_j^r + b^{(fl)}\right), \\
    f_{jr} & = & \sigma\left(U^{(fr)} h_j^l + b^{(fr)}\right), \\
    o_j & = & \sigma\left(U^{(o)} \coltwovec{h_j^l}{h_j^r} + b^{(o)}\right), \\
    u_j & = & \tanh\left(U^{(u)} \coltwovec{h_j^l}{h_j^r} + b^{(u)}\right), \\
    c_j & = & i_j \odot u_j + f_{jl} \odot c_{jl} + f_{jr} \odot c_{jr}, \\
    h_j & = & o_j \odot \tanh(c_j),
\end{eqnarray}
where the matrices $U \in R^{d \times 2d}$ (except for $U^{fl}$ and $U^{fr}$, for which $U \in R^{d \times d}$) and the biases $b \in R^{d}$ are the parameters to learn.
The memory state $c$ is controlled by $i$, $f$, and $o$ (called the input
gate, forget gate, and output gate, respectively), to hold important information for the entire network.
Each gate selectively activates to play a specific role (i.e., the input,
forget, and output gates control $u_j$, $c_{jl}$, and $c_{jr}$, respectively,
and $\tanh(c_j)$ is based on which elements should be input to the next state $c_j$, forgotten from the previous state $c_{jr}$ and $c_{jl}$, or output as a hidden representation $h_j$, respectively).

Note that the forget gate $f_{jl}$ for the left child state $c_{jl}$ only takes the right child's hidden representation $h_j^r$ and vice versa, as described by \newcite{Tai:2015}.

\subsection{Softmax Classifier with Attention}
We use a softmax classifier to predict the sentiment label $\hat{y}_j$ at any node $j$ for which a label is to be predicted.
Given the $j$th hidden representation $h_j$ as an input, the classifier predicts $\hat{y}_j$:
\begin{eqnarray}
    \hat{y}_j & = & \argmax_{y} \hat{p}_\theta(y|h_j), \\
    \hat{p}_\theta(y|h_j) & = & \softmax\left(W^{(s)}h_j + b^{(s)}\right),
\end{eqnarray}
where $W^{(s)} \in R^{d^l \times d}$ and $b^{(s)} \in R^n$ are the parameter matrix and bias vector for the classifier, respectively, and $d^l$ is the number of labels.
The softmax yields a label distribution $y \in R^{d^l}$, following which the classifier chooses the best label corresponding to the highest element among the $y$.

\begin{figure*}[t]
\begin{center}
\subfloat[RvNN]{
 \includegraphics[width=70mm]{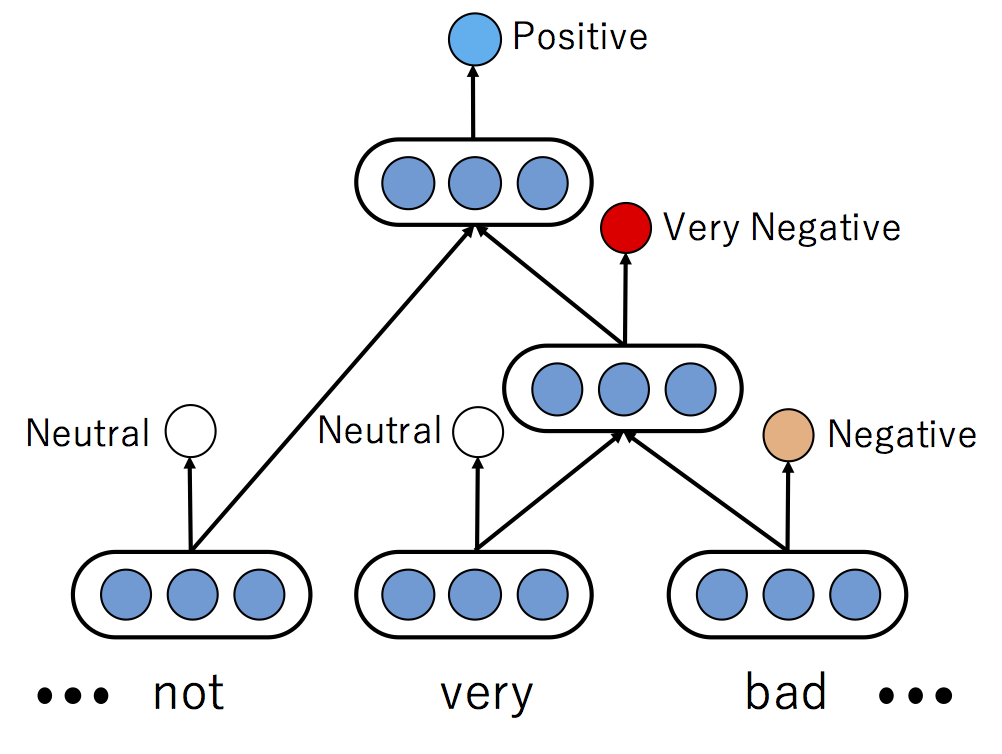}
 \label{fig:rnn}
}
\subfloat[RvNN with Attention]{
 \includegraphics[width=80mm]{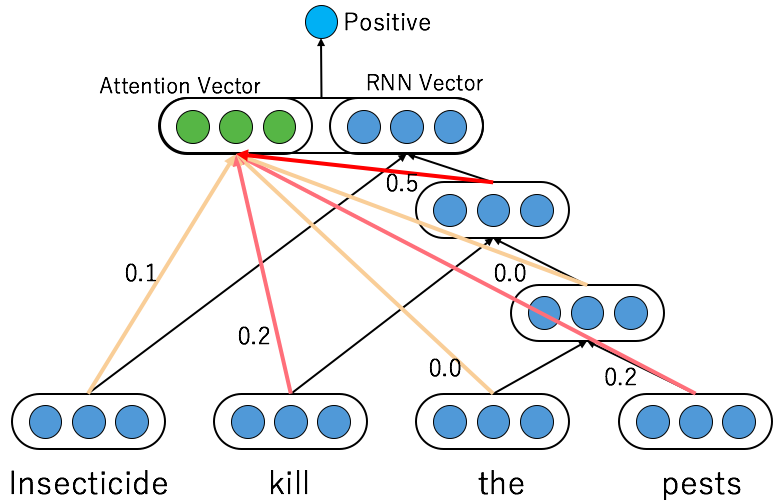}
}
\caption{Sentiment classification by Tree-LSTM with attention.}
\label{fig:attn_rnn}
\end{center}
\end{figure*}

\label{sect_attention}
However, owing to the lack of phrase-level annotation, sentence representation may be inaccurate because it may fail to propagate errors from the root of the tree to the terminals and pre-terminals in a long sentence.
We propose an attention mechanism to address this problem.
This so-called classifier with attention takes an attention vector representation $a_j$ in addition to a hidden representation $h_j$ as inputs:
\begin{eqnarray}
    \hat{p}_\theta(y|h_j)  &=&  \softmax\left(W^{(s^\prime)} \coltwovec{a_j}{h_j} + b^{(a)}\right), \\
    a_j & = & \sum_{i} a_{ji} \odot h_i, \\
    a_{ji} & = & \frac{g(h_i, h_j)}{\sum_{i^\prime}g(h_{i^\prime}, h_j)},  \\
    g(h_i, h_j) &=&  \exp\left(W^{(a2)}\tanh\left(W^{(a1)}\coltwovec{h_i}{h_j}\right)\right), \nonumber \\ \label{eq:attention3}
\end{eqnarray}
where $W^{(s^\prime)} \in R^{d^l \times 2d}, W^{(a1)} \in R^{d^a \times d}$, and $W^{(a2)} \in R^{1 \times d^a}$ are the parameter matrices.
In Eq.~\ref{eq:attention3}, the biases for both $W^{(a1)}$ and $W^{(a2)}$ are omitted for simplicity.
The attention vector $a_j$ represents how much the classifier pays attention to the children nodes of the target node.
The scalar values $a_{ji}$ for each node are used to determine the attention vector.

Figure \ref{fig:attn_rnn} represents the softmax classifier with attention.
\newcite{Kokkinos:2017} also investigated an attentional model for RvNNs.
But, their model only feeds the attention vector into the softmax classifier, whereas
our method inputs both the attention vector and RvNN vector, as illustrated in Figure \ref{fig:attn_rnn}.

\subsection{Distant Supervision with Polar Dictionaries}
\label{sect_dic}
Unlike the Stanford Sentiment Treebank, which is annotated with phrase-level polarity, other multilingual datasets contain only sentence-level annotation.
As shown in Section \ref{sect:experiments}, sentiment classification without a phrase-level annotated corpus will not learn sentence representations in an appropriate manner.
Although a phrase-level-polarity-tagged corpus is difficult to obtain in many languages, polar dictionaries are easy to compile (semi-)automatically.
Therefore, we opt for the use of polar dictionaries as an alternative source of sentiment information.

We utilize the same polar dictionaries for short phrases and words as used in \newcite{Nakagawa:2010}.
The phrase in the training sets that matches an entry in the polar dictionaries is annotated with the corresponding polarity.
The key difference from \newcite{Nakagawa:2010} is that we use polar
dictionaries as a hard label in a manner similar to distant supervision
\cite{Mintz:2009}. In contrast, the previous work used polar dictionaries 
as a soft label for an initial hidden variable in Tree-CRF.
\newcite{Teng:2016} also incorporated sentiment lexicons into an recurrent neural network model.
Their method predicts weights for each sentiment score of subjective words to predict a sentence label.
Our method uses polar dictionaries only during the training step, whereas the method by \newcite{Teng:2016} needs polar dictionaries for both training and decoding.

\subsection{Learning}
The cost function is a cross-entropy error function between the true class
label distribution, $t$ (i.e., one hot distribution for the correct label)
and the predicted label distribution, $\hat{y}$, at each labeled node:
\begin{equation}
    J(\theta) = -\sum^m_{k=1}t_k\log \hat{y}_k + \frac{\lambda}{2}\|\theta\|^2_2,
\end{equation}
where $m$ is the number of labeled nodes in the training set\footnote{If the
dataset contains only sentence-level annotation, $m$ is equal to the size of the dataset.}, and $\lambda$ denotes an L2 regularization hyperparameter.

\section{Experiments on Sentiment Classification}
\label{sect:experiments}
We conducted sentiment classification on a Japanese corpus where phrase-level
annotation is unavailable.
In addition, we performed sentiment classification on an English corpus
where phrase-level annotation is available, but without using phrase-level
annotation, to see its effect.

\subsection{Data}
\paragraph{Word embeddings.}
For Japanese experiments, we obtained pre-trained word representations from word2vec\footnote{See \url{https://code.google.com/archive/p/word2vec/}} using a skip-gram model \cite{Mikolov:2013a,Mikolov:2013b,Mikolov:2013c}.
We learned word representations on Japanese Wikipedia's dump data (2014.11) segmented by KyTea (version-0.4.7) \cite{Neubig:2011}.
We used pre-trained GloVe word
representations\footnote{See \url{http://nlp.stanford.edu/data/glove.6B.zip}} for the English experiments.
We fine-tuned both word representations in our experiments.

\paragraph{Parse trees.}
For Japanese constituency parsing, we used Ckylark \cite{Oda:2015} as of
2016.07\footnote{See \url{https://github.com/odashi/ckylark}} with KyTea for word segmentation. For English, we used the automatic syntactic annotation of the Stanford Sentiment Treebank.

\paragraph{Dictionaries.}
We followed \newcite{Nakagawa:2010} to create polar dictionaries.
We employed a Japanese polar dictionary composed by \newcite{Kobayashi:2005}
and
\newcite{Higashiyama:2008}\footnote{See \url{http://www.cl.ecei.tohoku.ac.jp/index.php?Open
Resources/Japanese Sentiment Polarity Dictionary}} that contains 5,447
positive and 8,117 negative expressions.\footnote{Note that these figures are
slightly different from \newcite{Nakagawa:2010}. We suspect that the reason
why they can use a larger lexicon is that they used an in-house (not publicly
available) version of the lexicon.} We created an English polar lexicon
from \newcite{Wilson:2005} in the same way as \newcite{Nakagawa:2010}.
The dictionary contains 2,289 positive and 4,143 negative
expressions.

\paragraph{Corpora.}
We used the NTCIR Japanese opinion corpus (NTCIR-J), which includes 997
positive and 2,400 negative sentences \cite{Seki:2007,Seki:2008}. We removed
neutral sentences following previous studies.
The corpus comprised two NTCIR Japanese opinion corpora, the NTCIR-6 corpus and the NTCIR-7 corpus, as in \cite{Nakagawa:2010}.
We performed 10-fold cross-validation by randomly splitting each corpus into
10 parts (one for testing, one for development, and the remaining eight for
training).\footnote{\newcite{Nakagawa:2010}
did not use development data to train the model, which means our model uses
only 86.6\% of the instances to train the model compared with theirs.}
For the English experiments, we used the Stanford Sentiment
Treebank \cite{Socher:2013}. It includes 11,855 sentences.
We followed the official training/development/testing split (8,544/1,101/2,210).
We used only sentence-level sentiment for our experiment.

\subsection{Methods}
In the Japanese experiments, we compared our method with seven baselines.
In the English experiments, we compared our method with two baselines.
All input word vectors, other than those for most frequent sentiment (MFS) and
Tree-CRF, were pre-trained by word2vec in Japanese experiments and by GloVe
in English experiments.
We implemented our method, LogRes, RvNN, Tree-LSTM, and our reimplementation of
\newcite{Kokkinos:2017} using Chainer \cite{Tokui:2015}.

The following methods were used.

\paragraph{MFS.}
A na\"{\i}ve baseline, as it always selects the most frequent sentiment (which is negative in this case).

\paragraph{LogRes.}
A linear classifier using logistic regression.
The input features are an average of word vectors in a sentence.

\paragraph{CNN}

The CNN-based sentiment classification
\cite{Kim:2014}.\footnote{See \url{https://github.com/yoonkim/CNN_sentence}}

\paragraph{Tree-CRF.}
A dependency-based tree-structured CRF \cite{Nakagawa:2010}.
This is the state-of-the-art method among our experimental
datasets.\footnote{Note that we cannot directly compare our result
with a stacked denoising auto-encoder \cite{Zhang:2015} (which
achieved slightly higher accuracy in the NTCIR-6 corpus,)
because we use a different dataset in our experiment.}

\paragraph{RvNN.}
The simplest RvNN.

\paragraph{Tree-LSTM.}

The LSTM-based RvNN \cite{Tai:2015}.

\paragraph{\newcite{Kokkinos:2017}}

Our reimplementation of \newcite{Kokkinos:2017}.
We implemented their TreeGRU model with LSTM instead of GRU.

\paragraph{Tree-LSTM w/ attn, dict.}
Our proposed method, which classifies polarity using attention and/or polar dictionaries.

\subsection{Hyperparameters}
The parameters used in both experiments are listed in Table \ref{table:hyper}.
For Japanese experiments, we tuned hyperparameters on each development set of 10-fold cross-validation.
For English experiments, we used similar hyperparameters with slight
modifications to the Japanese experiments.

\begin{table}[t]
\centering
 \begin{tabular}{lrr}
    \toprule
    \multicolumn{1}{c}{Parameter} & \multicolumn{2}{c}{Value} \\
     ~ & \multicolumn{1}{c}{Japanese} & \multicolumn{1}{c}{English}\\
    \midrule
    Word vector size   & 200 & 300 \\
    Hidden vector size & 200 & 200 \\
    Optimizer          & AdaDelta & AdaGrad \\
     \footnotesize{(Weight decay/learning rate)}  & 0.0001 & 0.005\\
    Gradient clipping & 5   & 5 \\
    \bottomrule
    \end{tabular}
    \caption{The hyperparameters.}
\label{table:hyper}
 \end{table}

\subsection{Results}
\begin{table}[t]
\centering
 \begin{tabular}{lr}
    \toprule
    \multicolumn{1}{c}{Method} & \multicolumn{1}{c}{Accuracy} \\
    \midrule
    MFS                    & 0.704  \\
    LogRes                 & 0.771\\
    CNN \cite{Kim:2014}    & 0.803\\
    Tree-CRF \cite{Nakagawa:2010} & 0.826\\
    \midrule
    RvNN                   & 0.517 \\
    Reimplementation of \newcite{Tai:2015} & 0.709\\
    Reimplementation of K\&P (2017) & 0.807\\
    \midrule
    Tree-LSTM w/ attn       & 0.810 \\
    Tree-LSTM w/ dict       & 0.829 \\
    Tree-LSTM w/ attn, dict & \bf{0.844} \\
    \bottomrule
    \end{tabular}
    \caption{Accuracy of each method on the Japanese sentiment classification task.}
\label{table:result1}

\end{table}
\begin{table}[t]
\centering
 \begin{tabular}{lr}
    \toprule
    \multicolumn{1}{c}{Method} & \multicolumn{1}{c}{Accuracy} \\
    \midrule
    \cite{Kim:2014}        & ~ \\
    --- CNN                & 48.0 \\
    \cite{Tai:2015}        & ~ \\
    --- Tree-LSTM          & 51.0 \\
    \cite{Kokkinos:2017}   & ~ \\
    --- TreeGRU w/o attn       & 50.5 \\
    --- TreeGRU w/ attn       & 51.0 \\
    \midrule
    Tree-LSTM               & 43.52 \\
    Tree-LSTM w/ attn       & 44.97 \\
    Tree-LSTM w/ dict       & 43.13 \\
    Tree-LSTM w/ attn, dict & 41.67 \\
    \bottomrule
    \end{tabular}
    \caption{Accuracy of each method on the English sentiment classification task.
    Note that our model learns only from sentence-level annotation.}
\label{table:result_en}

\end{table}

The Japanese experimental results are listed in Table \ref{table:result1}.
The accuracy of RvNN is much lower than that of the MFS baseline. Moreover,
Tree-LSTM, which is an improved RvNN, is still lower than simple LogRes,
despite Tree-LSTM achieving state-of-the-art performance on the
phrase-annotated Stanford Sentiment Treebank \cite{Tai:2015}.
The accuracy of CNN and \newcite{Kokkinos:2017} is 0.803 and 0.807,
respectively, which are slightly lower than that of our proposed method without polar dictionaries.
In contrast, Tree-LSTM with dictionary achieves comparable results to Tree-CRF.
Our Tree-LSTM with attention and polar dictionary obtained the highest accuracy.

The results of the English experiments are listed in Table \ref{table:result_en}.
Similarly to \newcite{Kokkinos:2017}, we observe slight improvement when using
attention for Tree-LSTM. However, unlike the Japanese experiments, using a dictionary degrades
sentiment classification accuracy. We discuss this in the following section.

\section{Discussion}
\label{sect:discussion}

\begin{figure*}[t]
\begin{center}
\subfloat[Consistency of policy cannot be found.]{
 \includegraphics[width=110mm]{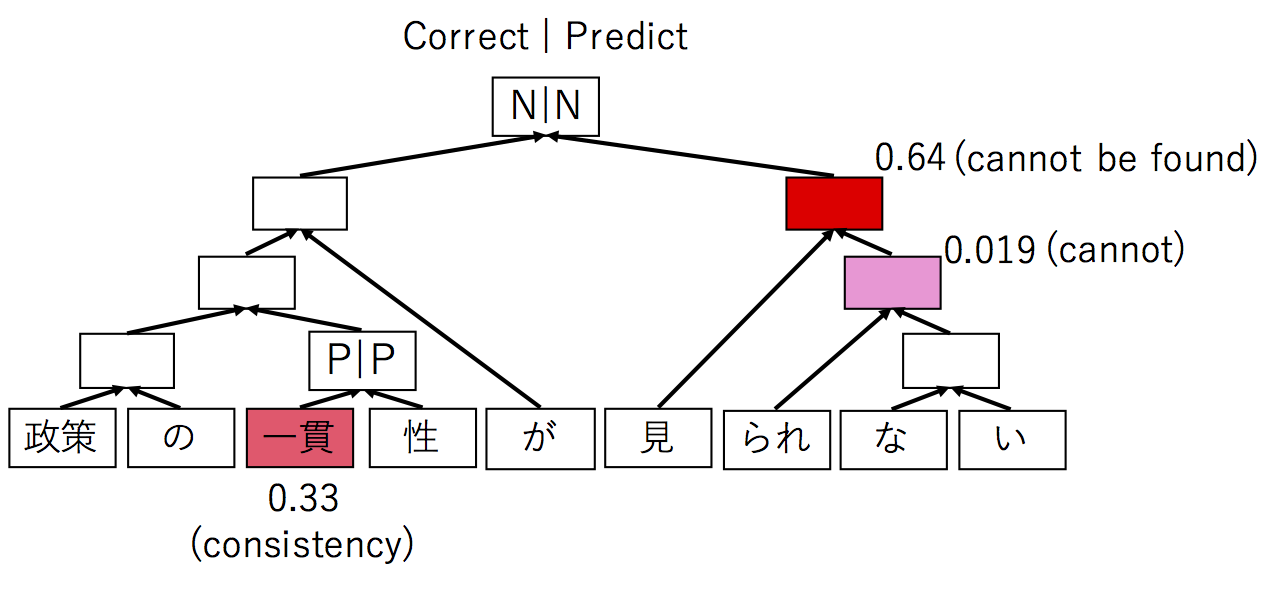}
 \label{fig:true1}
 }
\subfloat[Apprehension about friendship.]{
 \includegraphics[width=50mm]{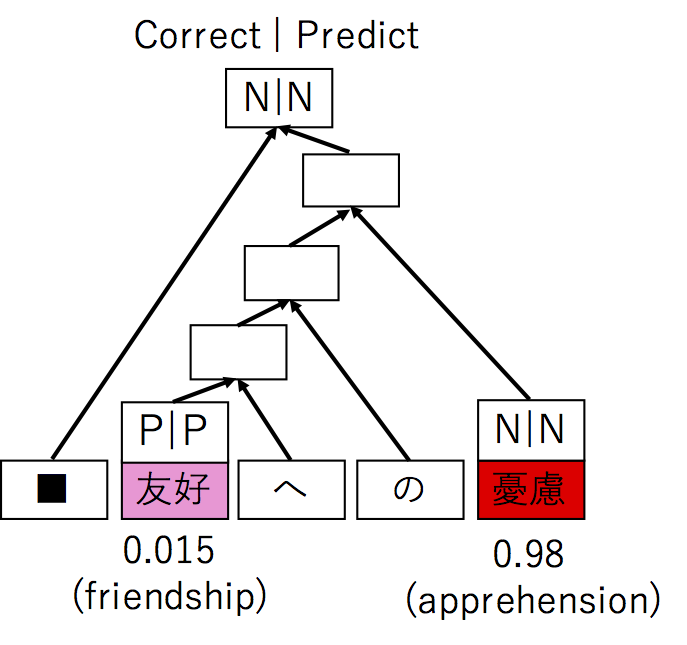}
 \label{fig:true2}
 } \\
\subfloat[Military confrontation was mitigated.]{
 \includegraphics[width=70mm]{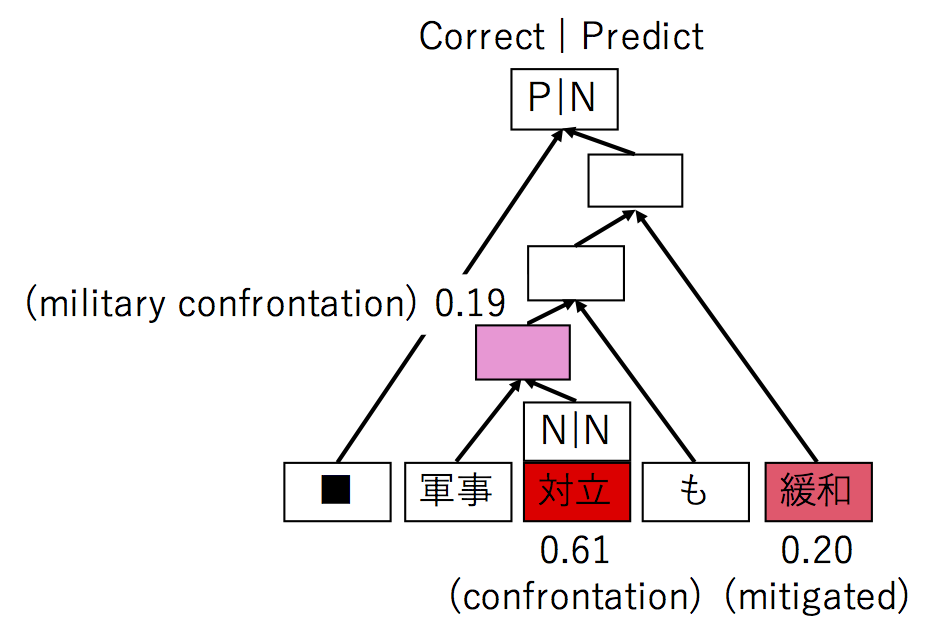}
 \label{fig:false1}
 }
\subfloat[Anyway, it was able to avoid the worst-case scenario.]{
 \includegraphics[width=90mm]{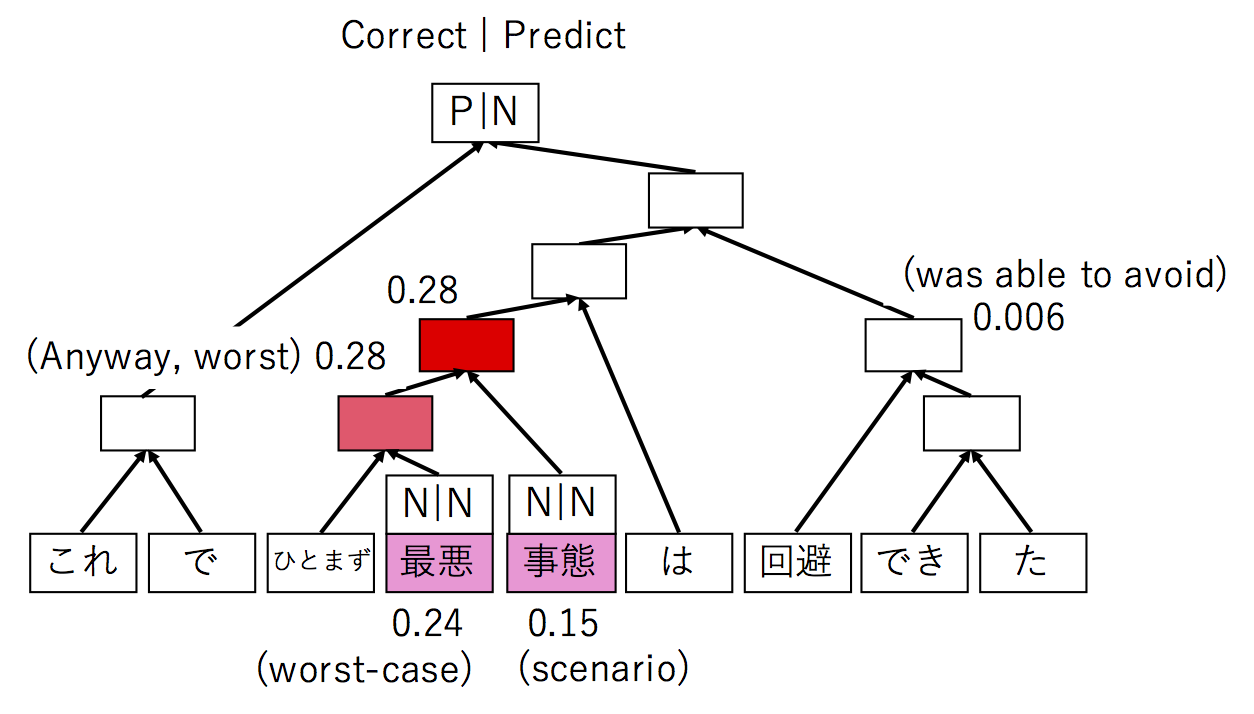}
 \label{fig:false2}
 }
\caption{Examples of our attentional Tree-LSTM sentiment classification on the test set. The red square indicates a word or phrase to which great attention was paid in the softmax step, and the associated value indicates the attention weight.
Root nodes are indicated by the (left) gold and (right) predicted labels (``N'' indicates negative, whereas ``P'' indicates positive).
We also show the labels for nodes that match an entry in the polar dictionary.
}
\label{fig:example}
\end{center}
\end{figure*}

\subsection{Effect of Attention and Dictionary}

The results described above indicate that Tree-LSTM models without an attention mechanism fail to learn sentence representations if phrase-level annotation is not available.

However, Tree-LSTM models can learn more accurate sentence representations if the models receive phrase-level information such as that provided by polar dictionaries.
For example, in our model, attention information and a polar dictionary are fed into the Tree-LSTM as phrase-level information.
Although Tree-LSTM with attention and a polar dictionary outperforms Tree-CRF
by 1.8 points, the accuracy of CNN, Tree-LSTM without a polar dictionary, and
\newcite{Kokkinos:2017} are lower than that of Tree-CRF.
Tree-LSTM with a polar dictionary performs better than Tree-LSTM with
attention, showing that a supervised label for each phrase seems to be important in learning Tree-LSTM models.

Table \ref{table:result_en} indicates that although an attention mechanism is
effective in both sentence-level and phrase-level annotated corpora,
the effect of dictionary information varies across datasets.
It is known that the effect of the English lexicon \cite{Wilson:2005}
is milder in the review dataset than in other domains \cite{Nakagawa:2010},
and we suppose that it introduces noise in the Stanford Sentiment Treebank.

\subsection{Examples}

Figures \ref{fig:true1} and \ref{fig:true2} show correctly classified examples.
Figure \ref{fig:true1} shows that the model classifies ``一貫性 (consistency)'' as positive and pays 1/3 attention to it in the final classification step; however, the model correctly classifies the sentence polarity as negative by considering ``見られない (cannot be found)'' through most of the attention.
In Figure \ref{fig:true2}, the model correctly classifies both ``友好 (friendship)'' and ``憂慮 (apprehension),'' and then classifies the sentence polarity by paying great attention to ``憂慮 (apprehension).''

Figures \ref{fig:false1} and \ref{fig:false2} display an incorrectly classified example.
In Figure \ref{fig:false1}, the model pays attention to both ``対立 (confrontation)'' and ``緩和 (mitigated)''; however, it fails to predict the correct polarity of the sentence.
It seems that the higher attention weight for ``対立 (confrontation)'' than for ``緩和 (mitigated)'' has an influence on the sentence prediction.
In Figure \ref{fig:false2}, the model fails to capture negation as it should pay attention to ``回避できた (was able to avoid).''
To solve these errors, the composition function should also incorporate an attention mechanism to handle polarity shifting correctly.

\section{Conclusion}
\label{sect:conclusion}
We have presented a Tree-LSTM-based RvNN using an attention mechanism and a polar dictionary.
In this method, each phrase representation is fed into a classifier to predict the polarity of a phrase based on the phrase structures.
Lexical items from the polar dictionary are used as supervised labels for each corresponding phrase or word in the same manner as distant supervision.
Our experimental results have demonstrated that the proposed method outperforms the
previous methods on a Japanese sentiment classification task.

Currently, we have annotated fine-grained phrase-level sentiment tags to a
Japanese review corpus. We plan to analyze the effect of phrase-level
annotation on Japanese sentiment analysis. In addition, we would like to
extend our attentional mechanism to use the Transformer \cite{Vaswani:2017} as proposed in \newcite{Shen:2018}.

\bibliographystyle{acl}
\bibliography{refs}

\begin{thebibliography}{}

\bibitem[\protect\citename{Bahdanau \bgroup et al.\egroup }2015]{Bahdanau:2015}
Dzmitry Bahdanau, KyungHyun Cho, and Yoshua Bengio.
\newblock 2015.
\newblock Neural machine translation by jointly learning to align and
  translate.
\newblock In {\em Proceedings of the 3rd International Conference on Learning
  Representations}.

\bibitem[\protect\citename{Eriguchi \bgroup et al.\egroup }2016]{Eriguchi:2016}
Akiko Eriguchi, Kazuma Hashimoto, and Yoshimasa Tsuruoka.
\newblock 2016.
\newblock Tree-to-sequence attentional neural machine translation.
\newblock In {\em Proceedings of the 54th Annual Meeting of the Association for
  Natural Language Processing}, page 823–833.

\bibitem[\protect\citename{Hermann \bgroup et al.\egroup }2015]{Hermann:2015}
Karl~Moritz Hermann, Tom{\'a}{\u{s}} Ko{\u{c}}isk{\'y}, Edward Grefenstette,
  Lasse Espeholt, Will Kay, Mustafa Suleyman, and Phil Blunsom.
\newblock 2015.
\newblock Teaching machines to read and comprehend.
\newblock In {\em Advances in Neural Information Processing Systems 28}, pages
  1693--1701.

\bibitem[\protect\citename{Higashiyama \bgroup et al.\egroup
  }2008]{Higashiyama:2008}
Masahiko Higashiyama, Kentaro Inui, and Yuji Matsumoto.
\newblock 2008.
\newblock Learning sentiment of nouns from selectional preferences of verbs and
  adjectives.
\newblock In {\em Proceedings of the 14th Annual Meeting of the Association for
  Natural Language Processing}, pages 584--587.

\bibitem[\protect\citename{Hochreiter and Schmidhuber}1997]{Hochreiter:1997}
Sepp Hochreiter and J{\"u}rgen Schmidhuber.
\newblock 1997.
\newblock Long short-term memory.
\newblock {\em Neural Computation}, 9(8):1735--1780.

\bibitem[\protect\citename{Kalchbrenner \bgroup et al.\egroup
  }2014]{Kalchbrenner:2014}
Nal Kalchbrenner, Edward Grefenstette, and Phil Blunsom.
\newblock 2014.
\newblock A convolutional neural network for modelling sentences.
\newblock In {\em Proceedings of the 52nd Annual Meeting of the Association for
  Natural Language Processing}, pages 655--665.

\bibitem[\protect\citename{Kim}2014]{Kim:2014}
Yoon Kim.
\newblock 2014.
\newblock Convolutional neural networks for sentence classification.
\newblock In {\em Proceedings of the 2014 Conference on Empirical Methods in
  Natural Language Processing}, pages 1746--1751.

\bibitem[\protect\citename{Kobayashi \bgroup et al.\egroup
  }2005]{Kobayashi:2005}
Nozomi Kobayashi, Kentaro Inui, Yuji Matsumoto, and Kenji Tateishi.
\newblock 2005.
\newblock Collecting evaluative expressions for opinion extraction.
\newblock {\em Journal of Natural Language Processing}, 12(3):203--222.

\bibitem[\protect\citename{Kokkinos and Potamianos}2017]{Kokkinos:2017}
Filippos Kokkinos and Alexandros Potamianos.
\newblock 2017.
\newblock Structural attention neural networks for improved sentiment analysis.
\newblock In {\em Proceedings of the 15th Conference of the European Chapter of
  the Association for Computational Linguistics}, pages 586--591.

\bibitem[\protect\citename{Ling \bgroup et al.\egroup }2015]{Ling:2015}
Wang Ling, Lin Chu-Cheng, Yulia Tsvetkov, Silvio Amir, Ram{\'o}n~Fernandez
  Astudillo, Chris Dyer, Alan~W Black, and Isabel Trancoso.
\newblock 2015.
\newblock Not all contexts are created equal: Better word representations with
  variable attention.
\newblock In {\em Proceedings of the 2015 Conference on Empirical Methods in
  Natural Language Processing}, pages 1367--1372.

\bibitem[\protect\citename{Luong \bgroup et al.\egroup }2015]{Luong:2015}
Minh-Thang Luong, Hieu Pham, and Christopher~D. Manning.
\newblock 2015.
\newblock Effective approaches to attention-based neural machine translation.
\newblock In {\em Proceedings of the 2015 Conference on Empirical Methods in
  Natural Language Processing}, pages 1412--1421.

\bibitem[\protect\citename{Mikolov \bgroup et al.\egroup }2013a]{Mikolov:2013a}
Tomas Mikolov, Kai Chen, Greg Corrado, and Jeffrey Dean.
\newblock 2013a.
\newblock Efficient estimation of word representations in vector space.
\newblock In {\em Proceedings of the 1st International Conference on Learning
  Representations}.

\bibitem[\protect\citename{Mikolov \bgroup et al.\egroup }2013b]{Mikolov:2013b}
Tomas Mikolov, Ilya Sutskever, Kai Chen, Greg Corrado, and Jeff Dean.
\newblock 2013b.
\newblock Distributed representations of words and phrases and their
  compositionality.
\newblock In {\em Advances in Neural Information Processing Systems 26}, pages
  3111--3119.

\bibitem[\protect\citename{Mikolov \bgroup et al.\egroup }2013c]{Mikolov:2013c}
Tomas Mikolov, Wen tau Yih, and Geoffrey Zweig.
\newblock 2013c.
\newblock Linguistic regularities in continuous space word representations.
\newblock In {\em Proceedings of Human Language Technologies: The 2013 Annual
  Conference of the North American Chapter of the Association for Computational
  Linguistics}, pages 746--751.

\bibitem[\protect\citename{Mintz \bgroup et al.\egroup }2009]{Mintz:2009}
Mike Mintz, Steven Bills, Rion Snow, and Dan jurafsky.
\newblock 2009.
\newblock Distant supervision for relation extraction without labeled data.
\newblock In {\em Proceedings of the Joint Conference of the 47th Annual
  Meeting of the ACL and the 4th International Joint Conference on Natural
  Language Processing of the AFNLP}, pages 1003--1011.

\bibitem[\protect\citename{Nakagawa \bgroup et al.\egroup }2010]{Nakagawa:2010}
Tetsuji Nakagawa, Kentaro Inui, and Sadao Kurohashi.
\newblock 2010.
\newblock Dependency tree-based sentiment classification using {CRFs} with
  hidden variables.
\newblock In {\em Proceedings of Human Language Technologies: The 2010 Annual
  Conference of the North American Chapter of the Association for Computational
  Linguistics}, pages 786--794.

\bibitem[\protect\citename{Neubig \bgroup et al.\egroup }2011]{Neubig:2011}
Graham Neubig, Yosuke Nakata, and Shinsuke Mori.
\newblock 2011.
\newblock Pointwise prediction for robust, adaptable {Japanese} morphological
  analysis.
\newblock In {\em Proceedings of The 49th Annual Meeting of the Association for
  Computational Linguistics: Human Language Technologies}, pages 529--533.

\bibitem[\protect\citename{Oda \bgroup et al.\egroup }2015]{Oda:2015}
Yusuke Oda, Graham Neubig, Sakriani Sakti, Tomoki Toda, and Satoshi Nakamura.
\newblock 2015.
\newblock Ckylark: A more robust {PCFG-LA} parser.
\newblock In {\em Proceedings of the 2015 Conference of the North American
  Chapter of the Association for Computational Linguistics: Demonstrations},
  pages 41--45.

\bibitem[\protect\citename{Pang \bgroup et al.\egroup }2002]{Pang:2002}
Bo~Pang, Lillian Lee, and Shivakumar Vaithyanathan.
\newblock 2002.
\newblock Thumbs up? sentiment classification using machine learning
  techniques.
\newblock In {\em Proceedings of the 2002 Conference on Empirical Methods in
  Natural Language Processing}, pages 79--86.

\bibitem[\protect\citename{Qian \bgroup et al.\egroup }2015]{Qian:2015}
Qiao Qian, Bo~Tian, Minlie Huang, Yang Liu, Xuan Zhu, and Xiaoyan Zhu.
\newblock 2015.
\newblock Learning tag embeddings and tag-specific composition functions in
  recursive neural network.
\newblock In {\em Proceedings of the 53rd Annual Meeting of the Association for
  Computational Linguistics and the 7th International Joint Conference on
  Natural Language Processing}, pages 1365--1374.

\bibitem[\protect\citename{Rensink}2000]{Rensink:2000}
Ronald~A. Rensink.
\newblock 2000.
\newblock The dynamic representation of scenes.
\newblock {\em Visual cognition}, 7(1-3):17--42.

\bibitem[\protect\citename{Rush \bgroup et al.\egroup }2015]{Rush:2015}
Alexander~M. Rush, Sumit Chopra, and Jason Weston.
\newblock 2015.
\newblock A neural attention model for sentence summarization.
\newblock In {\em Proceedings of the 2015 Conference on Empirical Methods in
  Natural Language Processing}, pages 379--389.

\bibitem[\protect\citename{Seki \bgroup et al.\egroup }2007]{Seki:2007}
Yohei Seki, David~Kirk Evans, and Lun-Wei Ku.
\newblock 2007.
\newblock Overview of opinion analysis pilot task at {NTCIR}-6.
\newblock In {\em Proceedings of the 6th NTCIR Workshop}, pages 265--278.

\bibitem[\protect\citename{Seki \bgroup et al.\egroup }2008]{Seki:2008}
Yohei Seki, David~Kirk Evans, and Lun-Wei Ku.
\newblock 2008.
\newblock Overview of multilingual opinion analysis task at {NTCIR}-7.
\newblock In {\em Proceedings of the 7th NTCIR Workshop}, pages 265--278.

\bibitem[\protect\citename{Shen \bgroup et al.\egroup }2018]{Shen:2018}
Tao Shen, Tianyi Zhou, Guodong Long, Jing Jiang, Shirui Pan, and Chengqi Zhang.
\newblock 2018.
\newblock Disan: Directional self-attention network for rnn/cnn-free language
  understanding.
\newblock In {\em Proceedings of the Thirty-Second {AAAI} Conference on
  Artificial Intelligence}, pages 5446--5455.

\bibitem[\protect\citename{Socher \bgroup et al.\egroup }2011]{Socher:2011}
Richard Socher, Jeffrey Pennington, Eric~H. Huang, Andrew~Y. Ng, and
  Christopher~D. Manning.
\newblock 2011.
\newblock Semi-supervised recursive autoencoders for predicting sentiment
  distributions.
\newblock In {\em Proceedings of the 2011 Conference on Empirical Methods in
  Natural Language Processing}, pages 151--161.

\bibitem[\protect\citename{Socher \bgroup et al.\egroup }2012]{Socher:2012}
Richard Socher, Brody Huvaland Christopher~D. Manning, and Andrew~Y. Ng.
\newblock 2012.
\newblock Semantic compositionality through recursive matrix-vector spaces.
\newblock In {\em Proceedings of the 2012 Joint Conference on Empirical Methods
  in Natural Language Processing and Computational Natural Language Learning},
  pages 1201--1211.

\bibitem[\protect\citename{Socher \bgroup et al.\egroup }2013]{Socher:2013}
Richard Socher, Alex Perelygin, Jean~Y. Wu, Jason Chuang, Christopher~D.
  Manning, Andrew~Y. Ng, and Christopher Potts.
\newblock 2013.
\newblock Recursive deep models for semantic compositionality over a sentiment
  treebank.
\newblock In {\em Proceedings of the 2013 Conference on Empirical Methods in
  Natural Language Processing}, pages 1631--1642.

\bibitem[\protect\citename{Tai \bgroup et al.\egroup }2015]{Tai:2015}
Kai~Sheng Tai, Richard Socher, and Christopher~D. Manning.
\newblock 2015.
\newblock Improved semantic representations from tree-structured long
  short-term memory networks.
\newblock In {\em Proceedings of the 53rd Annual Meeting of the Association for
  Computational Linguistics and the 7th International Joint Conference on
  Natural Language Processing}, pages 1556--1566.

\bibitem[\protect\citename{Teng \bgroup et al.\egroup }2016]{Teng:2016}
Zhiyang Teng, Duy-Tin Vo, and Yue Zhang.
\newblock 2016.
\newblock Context-sensitive lexicon features for neural sentiment analysis.
\newblock In {\em Proceedings of the 2016 Conference on Empirical Methods in
  Natural Language Processing}, pages 1629--1638.

\bibitem[\protect\citename{Tokui \bgroup et al.\egroup }2015]{Tokui:2015}
Seiya Tokui, Kenta Oono, Shohei Hido, and Justin Clayton.
\newblock 2015.
\newblock Chainer: a next-generation open source framework for deep learning.
\newblock In {\em Proceedings of Workshop on Machine Learning Systems in The
  29th Annual Conference on Neural Information Processing Systems}.

\bibitem[\protect\citename{Vaswani \bgroup et al.\egroup }2017]{Vaswani:2017}
Ashish Vaswani, Noam Shazeer, Niki Parmar, Jakob Uszkoreit, Llion Jones,
  Aidan~N Gomez, \L~ukasz Kaiser, and Illia Polosukhin.
\newblock 2017.
\newblock Attention is all you need.
\newblock In {\em Advances in Neural Information Processing Systems 30}, pages
  5998--6008.

\bibitem[\protect\citename{Vinyals \bgroup et al.\egroup }2015]{Vinyals:2015}
Oriol Vinyals, Meire Fortunato, and Navdeep Jaitly.
\newblock 2015.
\newblock Pointer networks.
\newblock In {\em Advances in Neural Information Processing Systems 28}, pages
  2692--2700.

\bibitem[\protect\citename{Wilson \bgroup et al.\egroup }2005]{Wilson:2005}
Theresa Wilson, Janyce Wiebe, and Paul Hoffmann.
\newblock 2005.
\newblock Recognizing contextual polarity in phrase- level sentiment analysis.
\newblock In {\em Proceedings of the 2005 Joint Conference on Human Language
  Technology and Empirical Methods in Natural Language Processing}, pages
  347--354.

\bibitem[\protect\citename{Xu \bgroup et al.\egroup }2015]{Xu:2015}
Kelvin Xu, Jimmy Ba, Ryan Kiros, Kyunghyun Cho, Aaron Courville, Ruslan
  Salakhudinov, Rich Zemel, and Yoshua Bengio.
\newblock 2015.
\newblock Show, attend and tell: Neural image caption generation with visual
  attention.
\newblock In {\em Proceedings of the 32nd International Conference on Machine
  Learning}, pages 2048--2057.

\bibitem[\protect\citename{Yang \bgroup et al.\egroup }2016]{Yang:2016}
Zichao Yang, Diyu Yang, Chris Dyer, Xiaodong He, Alex Smola, and Eduard Hovy.
\newblock 2016.
\newblock Hierarchical attention networks for document classification.
\newblock In {\em Proceedings of The 15th Annual Conference of the North
  American Chapter of the Association for Computational Linguistics: Human
  Language Technologies ({NAACL-HLT 2016})}, pages 1480--1489.

\bibitem[\protect\citename{Zhang and Komachi}2015]{Zhang:2015}
Peinan Zhang and Mamoru Komachi.
\newblock 2015.
\newblock Japanese sentiment classification with stacked denoising auto-encoder
  using distributed word representation.
\newblock In {\em Proceedings of the 29th Pacific Asia Conference on Language,
  Information and Computation}, pages 150--159.

\bibitem[\protect\citename{Zhu and Sobhani}2015]{Zhu:2015}
Xiaodan Zhu and Parinaz Sobhani.
\newblock 2015.
\newblock Long short-term memory over recursive structures.
\newblock In {\em Proceedings of the 32nd International Conference on Machine
  Learning}, pages 1604--1612.

\bibitem[\protect\citename{Zou \bgroup et al.\egroup }2018]{Zou:2018}
Yicheng Zou, Tao Gui, Qi~Zhang, and Xuanjing Huang.
\newblock 2018.
\newblock A lexicon-based supervised attention model for neural sentiment
  analysis.
\newblock In {\em Proceedings of the 27th International Conference on
  Computational Linguistics ({COLING})}, pages 868--877.

\end{thebibliography}

\end{document}